\documentclass[11pt,letterpaper]{article}
\usepackage{emnlp2016}
\usepackage{times}
\usepackage{latexsym}
\usepackage{xcolor}
\emnlpfinalcopy

\def\emnlppaperid{502}

\usepackage[utf8]{inputenc} 
\usepackage[T1]{fontenc}    
\usepackage{hyperref}
\definecolor{mydarkblue}{rgb}{0,0.08,0.45}
\hypersetup{
colorlinks,
citecolor=mydarkblue,
linkcolor=mydarkblue,
urlcolor=mydarkblue}       
\usepackage{url}            
\usepackage{booktabs}       
\usepackage{amsfonts}       
\usepackage{nicefrac}       
\usepackage{microtype}      

\usepackage{times}
\usepackage{latexsym}
\usepackage{amsmath,amssymb,enumerate,bbm,color,alltt}
\usepackage{hyperref}
\usepackage{url}
\usepackage{graphicx}
\usepackage{float}
\usepackage{stfloats}
\usepackage{subfig}
\usepackage{epstopdf}
\usepackage{longtable}
\usepackage{tabularx}
\usepackage{multirow}
\usepackage{booktabs}
\usepackage{subfig}
\usepackage[titletoc]{appendix}

\newcommand\Bstrut{\rule[-1.2ex]{0pt}{0pt}}   

\newcommand{\nin}{\not\in}
\newcommand{\nobracket}{}
\newcommand{\tmmathbf}[1]{\ensuremath{\boldsymbol{#1}}}
\newcommand{\tmop}[1]{\ensuremath{\operatorname{#1}}}

\newcommand{\cev}[1]{\reflectbox{\ensuremath{\vec{\reflectbox{\ensuremath{#1}}}}}}

\title{Language as a Latent Variable: \\
Discrete Generative Models for Sentence Compression}

\author{
  Yishu Miao$^1$, Phil Blunsom$^{1,2}$ \\
  $^1$University of Oxford, $^2$Google Deepmind\\
  \texttt{\{yishu.miao, phil.blunsom\}@cs.ox.ac.uk} \\
}

\begin{document}

\maketitle

\begin{abstract}
In this work we explore deep generative models of text in which the latent representation of a document is itself drawn from a discrete language model distribution. We formulate a variational auto-encoder for inference in this model and apply it to the task of compressing sentences. In this application the generative model first draws a latent summary sentence from a background language model, and then subsequently draws the observed sentence conditioned on this latent summary. In our empirical evaluation we show that generative formulations of both abstractive and extractive compression yield state-of-the-art results when trained on a large amount of supervised data. Further, we explore semi-supervised compression scenarios where we show that it is possible to achieve performance competitive with previously proposed supervised models while training on a fraction of the supervised data.
\end{abstract}

\section{Introduction}
The recurrent sequence-to-sequence paradigm for natural language generation \cite{kalchbrenner2013recurrent,sutskever2014sequence} has achieved remarkable recent success and is now the approach of choice for applications such as machine translation \cite{bahdanau2014neural}, caption generation \cite{xu2015show} and speech recognition \cite{chorowski2015attention}.
While these models have developed sophisticated conditioning mechanisms, e.g.\ attention, fundamentally they are discriminative models trained only to approximate the conditional output distribution of strings.
In this paper we explore modelling the joint distribution of string pairs using a deep generative model and employing a discrete variational auto-encoder (VAE) for inference \cite{kingma2013auto,rezende2014stochastic,mnih2014neural}. 
We evaluate our generative approach on the task of sentence compression.
This approach provides both alternative supervised objective functions and the opportunity to perform semi-supervised learning by exploiting the VAEs ability to marginalise the latent compressed text for unlabelled data. 

Auto-encoders \cite{rumelhart1985learning} are a typical neural network architecture for learning compact data representations, with the general aim of performing dimensionality reduction on embeddings \cite{hinton2006reducing}.
In this paper, rather than seeking to embed inputs as points in a vector space, we describe them with explicit natural language sentences.
This approach is a natural fit for summarisation tasks such as sentence compression.
According to this, we propose a generative  \textbf{auto-encoding sentence compression} (ASC) model, where we introduce a latent language model to provide the variable-length compact summary. 
The objective is to perform Bayesian inference for the posterior distribution of summaries conditioned on the observed utterances.
Hence, in the framework of  VAE, we construct an inference network as the variational approximation of the posterior, which generates compression samples to optimise the variational lower bound.

\begin{figure*}[!tbh]
  \centering
	\includegraphics[width=6in]{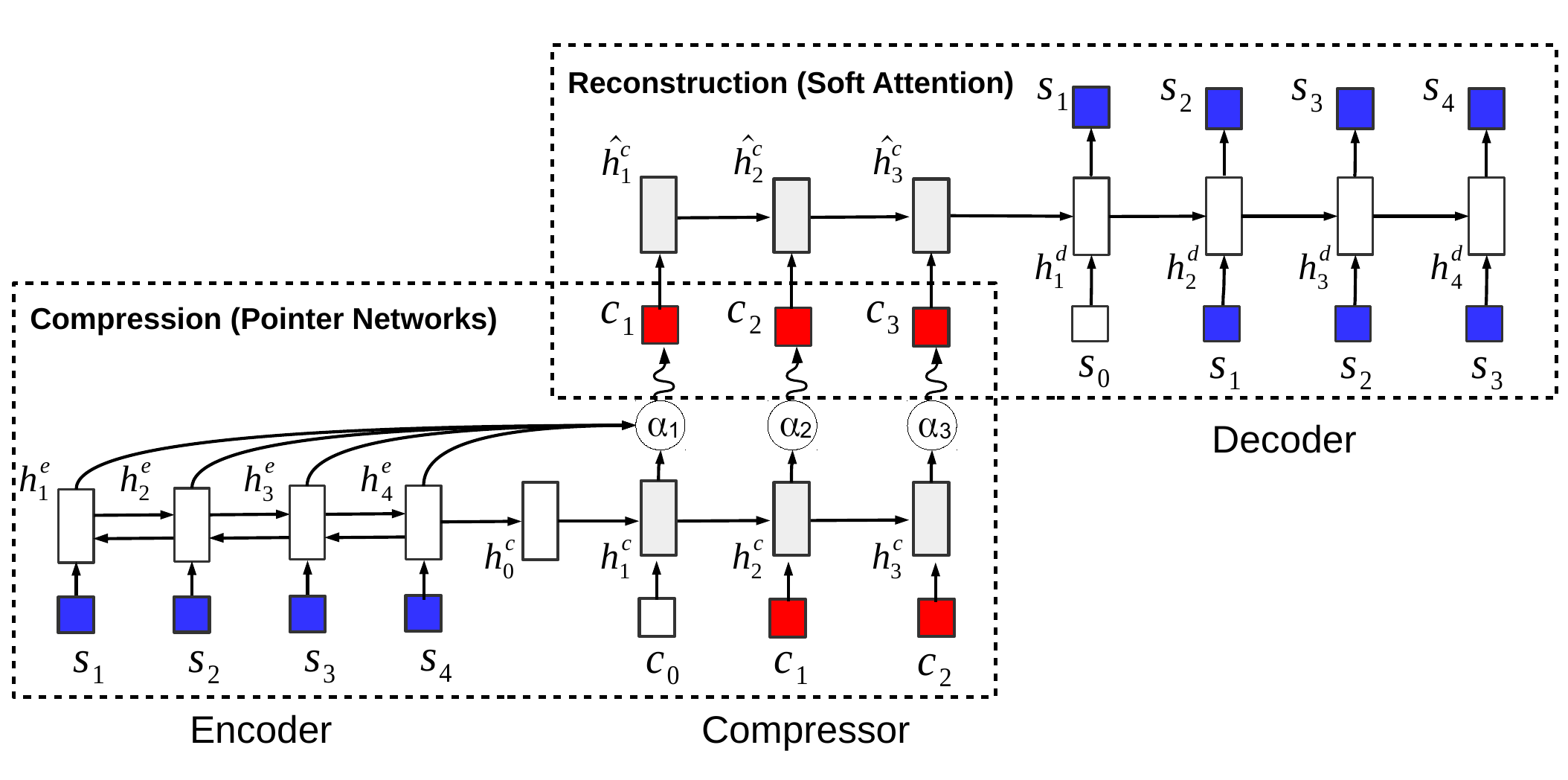} 
  \caption{Auto-encoding Sentence Compression Model}
  \label{fig:aec}
  	\vspace{-0.5em}
\end{figure*}

The most common family of variational auto-encoders relies on the reparameterisation trick, which is not applicable for our discrete latent language model. 
Instead, we employ the REINFORCE algorithm  \cite{mnih2014recurrent,mnih2014neural} to mitigate the problem of high variance during sampling-based variational inference.
Nevertheless, when directly applying the RNN encoder-decoder to model the variational distribution it is very difficult to generate reasonable compression samples in the early stages of training, since each hidden state of the sequence would have $|V|$ possible words to be sampled from.
To combat this we employ pointer networks \cite{vinyals2015pointer} to construct the variational distribution. This biases the latent space to sequences composed of words only appearing in the source sentence (i.e. the size of softmax output for each state becomes the length of current source sentence), which amounts to applying an extractive compression model for the variational approximation.

In order to further boost the performance on sentence compression, we employ a supervised \textbf{forced-attention sentence compression} model (FSC) trained on labelled data to teach the ASC model to generate compression sentences.
The FSC model shares the pointer network of the ASC model and combines a softmax output layer over the whole vocabulary.
Therefore, while training on the sentence-compression pairs, it is able to balance copying a word from the source sentence with generating it from the background distribution.
More importantly, by jointly training on the labelled and unlabelled datasets, this shared pointer network enables the model to work in a semi-supervised scenario. 
In this case, the FSC teaches the ASC to generate reasonable samples, while the pointer network trained on a large unlabelled data set helps the FSC model to perform better abstractive summarisation.

In Section 6, we evaluate the proposed model by jointly training the generative (ASC) and discriminative (FSC) models on the standard Gigaword sentence compression task with varying amounts of labelled and unlabelled data. The results demonstrate that by introducing a latent language variable we are able to match the previous benchmakers with small amount of the supervised data. When we employ our mixed discriminative and generative objective with all of the supervised data the model significantly outperforms all previously published results.

\section{Auto-Encoding Sentence Compression}
In this section, we introduce the auto-encoding sentence compression model (Figure \ref{fig:aec})\footnote{The language model, layer connections and decoder soft attentions are omitted in Figure \ref{fig:aec} for clarity.} in the framework of variational auto-encoders. The ASC model consists of four recurrent neural networks -- an encoder, a compressor, a decoder and a language model.

Let $\tmmathbf{s}$ be the source sentence, and $\tmmathbf{c}$ be the compression sentence. 
The \textbf{compression} model (encoder-compressor) is the inference network $q_{\phi}(\tmmathbf{c}|\tmmathbf{s})$ that takes source sentences $\tmmathbf{s}$ as inputs and generates extractive compressions $\tmmathbf{c}$.
The \textbf{reconstruction} model (compressor-decoder) is the generative network $p_{{\theta}}(\tmmathbf{s}|\tmmathbf{c})$ that reconstructs source sentences $\tmmathbf{s}$ based on the latent compressions $\tmmathbf{c}$.
Hence, the forward pass starts from the encoder to the compressor and ends at the decoder.
As the prior distribution, a language model $p(\tmmathbf{c})$ is pre-trained to regularise the latent compressions so that the samples drawn from the compression model are likely to be reasonable natural language sentences.

\subsection{Compression}

For the compression model (encoder-compressor), $q _{\phi}(\tmmathbf{c}|\tmmathbf{s})$, we employ a pointer network consisting of a bidirectional LSTM encoder that processes the source sentences, and an LSTM compressor that generates compressed sentences by attending to the encoded source words.

Let $\tmmathbf{s}_i$ be the words in the source sentences, $\tmmathbf{h}^e_{i}$ be the corresponding state outputs of the encoder.
$\tmmathbf{h}^e_{i}$ are the concatenated hidden states from each direction:
\begin{equation}
	\tmmathbf{h}^e_i  =  f_{\overrightarrow{\tmop{enc}}}(\vec{\tmmathbf{h}}\hspace{1pt}^e_{i-1}, \tmmathbf{s}_i)| |
	f_{\overleftarrow{\tmop{enc}}}(\cev{\tmmathbf{h}}\hspace{1pt}^e_{i+1}, \tmmathbf{s}_i)
	\label{eq:bi}
\end{equation}
Further, let $\tmmathbf{c}_j$ be the words in the compressed sentences, $\tmmathbf{h}^c_j$ be the state outputs of the compressor. We construct the predictive distribution by attending to the words in the source sentences:
\begin{align}
  \tmmathbf{h}^c_j  = & f_{\tmop{com}} ( \tmmathbf{h}^c_{j - 1},
  \tmmathbf{c}_{j-1}) \\
    \tmmathbf{u}_j(i)  = &   \tmmathbf{w}_3^T \! \tanh (\! \tmmathbf{W}\!\!_1 
  \tmmathbf{h}^c_j \!+\! \tmmathbf{W}\!_2  \tmmathbf{h}^e_i) \\
  \label{eq:u}
  q_\phi(\tmmathbf{c}_j |\tmmathbf{c}_{1:j\!-\!1},\tmmathbf{s}) \! = & \tmop{softmax} (\tmmathbf{u}_j)
\end{align}
where $\tmmathbf{c}_0$ is the start symbol for each compressed sentence and $\tmmathbf{h}^c_0$ is initialised by the source sentence vector of $\tmmathbf{h}^e_{|s|}$. 
In this case, all the words $\tmmathbf{c}_j$ sampled from $q_\phi(\tmmathbf{c}_j|\tmmathbf{c}_{1:j-1},\tmmathbf{s})$ are the subset of the words appeared in the source sentence (i.e. $\tmmathbf{c}_j\in\tmmathbf{s}$).

\subsection{Reconstruction}
For the reconstruction model (compressor-decoder) $p _{\theta}(\tmmathbf{s}|\tmmathbf{c})$, we apply a soft attention sequence-to-sequence model to generate the source sentence $\tmmathbf{s}$ based on the compression samples $\tmmathbf{c} \sim q_{\phi}(\tmmathbf{c}|\tmmathbf{s})$.

Let $\tmmathbf{s}_k$ be the words in the reconstructed sentences and $\tmmathbf{h}^d_{k}$ be the corresponding state outputs of the decoder:
\begin{equation}
  \tmmathbf{h}^d_k  =  f_{\tmop{dec}} ( \tmmathbf{h}^d_{k - 1},
  \tmmathbf{s}_{k-1})
\end{equation}
In this model, we directly use the recurrent cell of the compressor to encode the compression samples\footnote{The recurrent parameters of the compressor are not updated by the gradients from the reconstruction model.}:
\begin{align}
  \tmmathbf{\hat{h}}^c_j  = & f_{\tmop{com}} ( \tmmathbf{\hat{h}}^c_{j - 1}, \tmmathbf{c}_j)
\end{align}
where the state outputs $\tmmathbf{\hat{h}}^c_j$ corresponding to the word inputs $\tmmathbf{c}_j$ are different from the outputs $\tmmathbf{h}^c_j$ in the compression model, since we block the information from the source sentences.
We also introduce a start symbol $\tmmathbf{s}_0$ for the reconstructed sentence and $\tmmathbf{h}^d_0$ is initialised by the last state output $\tmmathbf{\hat{h}}^c_{|\tmmathbf{c}|}$. The soft attention model is defined as:
\begin{align}
  v_k(j)  = &   \tmmathbf{w}_6^T \tanh ( \tmmathbf{W}_4 
  \tmmathbf{h}^d_k + \tmmathbf{W}_5  \tmmathbf{\hat{h}}^c_j) \\
  \gamma_k(j)  = & \tmop{softmax} ( v_k(j)) \\
  \tmmathbf{d}_k = & \sum\nolimits_j^{\tmmathbf{|c|}} \gamma_k(j) \tmmathbf{\hat{h}}^c_j  ( v_k(j))
\end{align}
We then construct the predictive probability distribution over reconstructed words using a softmax:
\begin{align}
  & p_\theta ( \tmmathbf{s}_k | \tmmathbf{s}_{1:k-1}, \tmmathbf{c} ) = \tmop{softmax} ( \tmmathbf{W}_7 \tmmathbf{d}_k )
\end{align}

\begin{figure*}[!tb]
 \vspace{-1em}
  \centering
	\includegraphics[width=4.4in]{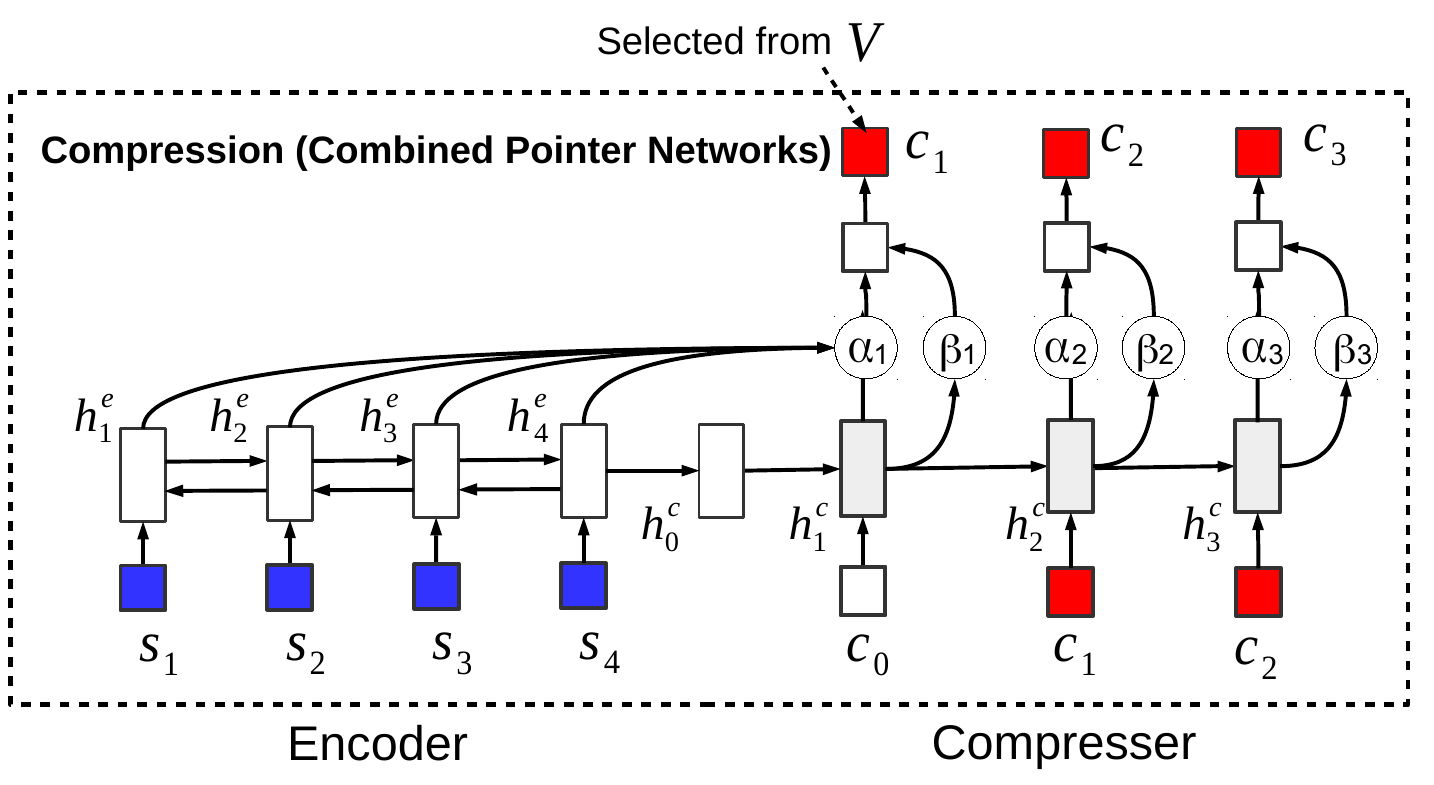} 
  \caption{Forced Attention Sentence Compression Model}
  \label{fig:fsc}
  \vspace{-1em}
\end{figure*}

\subsection{Inference}
In the ASC model there are two sets of parameters, $\phi$ and $\theta$, that need to be updated during inference. 
Due to the non-differentiability of the model, the reparameterisation trick of the VAE is not applicable in this case. 
Thus, we use the REINFORCE algorithm \cite{mnih2014recurrent,mnih2014neural} to reduce the variance of the gradient estimator.

The variational lower bound of the ASC model is:
\vspace{-0.5em}
\begin{align}
   \nonumber L = &  \mathbbm{E} _{q_{\phi}(\tmmathbf{c}|\tmmathbf{s})}[\log p_\theta(\tmmathbf{s}|\tmmathbf{c})]-D_{KL}[ q_{\phi}(\tmmathbf{c}|\tmmathbf{s})|| p(\tmmathbf{c})] \\
  \leqslant & \log \int \!\! \frac{q_{\phi}(\tmmathbf{c}|\tmmathbf{s})}{q_{\phi}(\tmmathbf{c}|\tmmathbf{s})} p_\theta(\tmmathbf{s}|\tmmathbf{c})  p(\tmmathbf{c}) d \tmmathbf{c} = \log p(\tmmathbf{s})
\label{eq:lb}
\end{align}
Therefore, by optimising the lower bound (Eq. \ref{eq:lb}), 
the model balances the selection of keywords for the summaries and the efficacy of the composed compressions, corresponding to the reconstruction error and KL divergence respectively.

In practise, the pre-trained language model prior $p(\tmmathbf{c})$ prefers short sentences for compressions.
As one of the drawbacks of VAEs, the KL divergence term in the lower bound pushes every sample drawn from the variational distribution towards the prior. 
Thus acting to regularise the posterior, but also to restrict the learning of the encoder.
If the estimator keeps sampling short compressions during inference, the LSTM decoder would gradually rely on the contexts from the decoded words instead of the information provided by the compressions, which does not yield the best performance on sentence compression.

Here, we introduce a co-efficient $\lambda$ to scale the learning signal of the KL divergence:
\vspace{-0.5em}
\begin{align}
\! \!    L \!\!  = &  \mathbbm{E} _{q_{\phi}(\tmmathbf{c}|\tmmathbf{s})}[\log p_\theta(\tmmathbf{s}|\tmmathbf{c})]\!\!  -\!\!  \lambda D\!_{K\!L}[ q_{\phi}( \!\tmmathbf{c}|\tmmathbf{s})|| p(\! \tmmathbf{c})] 
\label{eq:scale}
\end{align}
Although we are not optimising the exact variational lower bound, the ultimate goal of learning an effective compression model is mostly up to the reconstruction error.
In Section 6, we empirically apply $\lambda=0.1$ for all the experiments on ASC model. Interestingly, $\lambda$ controls the compression rate of the sentences which can be a good point to be explored in future work.

During the inference, we have different strategies for updating the parameters of $\phi$ and $\theta$. For the parameters $\theta$ in the reconstruction model, we directly update them by the gradients:
\begin{align}
\frac{\partial L}{\partial \theta}   & = \mathbbm{E} _{q_{\phi}(\tmmathbf{c}|\tmmathbf{s})}[\frac{\partial \log p_{\theta} ( \tmmathbf{s} | \tmmathbf{c})}{\partial \theta}]  \nonumber \\
 & \approx  \frac{1}{M}  \sum_m
  \frac{\partial \log p_{\theta} ( \tmmathbf{s} | \tmmathbf{c}
  ^{(m)})}{\partial \theta}
\end{align}
where we draw $M$ samples $\tmmathbf{c}^{(m)} \sim q_{\phi}(\tmmathbf{c}|\tmmathbf{s})$ independently for computing the stochastic gradients.

For the parameters $\phi$ in the compression model,  we firstly define the learning signal,
\begin{align}
\nonumber l(\tmmathbf{s}, \tmmathbf{c}) & =  \log p_\theta ( \tmmathbf{s} | \tmmathbf{c}) - \lambda (\log  q_\phi ( \tmmathbf{c} | \tmmathbf{s}) - \log p ( \tmmathbf{c})  ). 
\end{align}
Then, we update the parameters $\phi$ by:
\begin{align}
 \frac{\partial L}{\partial \phi} & = \mathbbm{E} _{q_{\phi}(\tmmathbf{c}|\tmmathbf{s})}[l(\tmmathbf{s}, \tmmathbf{c})\frac{\partial \log q_\phi (\tmmathbf{c} | \tmmathbf{s} \nobracket)}{\partial \phi}]  \nonumber \\
 &\approx  \frac{1}{M}  \sum_m [l(\tmmathbf{s}, \tmmathbf{c}^{(m)}) \frac{\partial \log q_\phi (\tmmathbf{c}^{(m\!)} | \tmmathbf{s} \nobracket)}{\partial \phi}]
\end{align}
However, this gradient estimator has a big variance because the learning signal $l(\tmmathbf{s}, \tmmathbf{c}^{(m)})$ relies on the samples from $q_{\phi}(\tmmathbf{c}|\tmmathbf{s})$. Therefore, following the REINFORCE algorithm, we introduce two baselines $b$ and $b(\tmmathbf{s})$, the centred learning signal and input-dependent baseline respectively, to help reduce the variance. 

Here, we build an MLP to implement the input-dependent baseline $b(\tmmathbf{s})$.
During training, we learn the two baselines by minimising the expectation:
\begin{align}
	&\mathbbm{E} _{q_{\phi}( \tmmathbf{c} | \tmmathbf{s} )}[ (l(\tmmathbf{s}, \tmmathbf{c})-b-b(\tmmathbf{s}))^2].
\end{align}
Hence, the gradients w.r.t. $\phi$ are derived as,
\begin{equation}
 \frac{\partial L}{\partial \phi} \! \approx \!\!  \frac{1}{M} \!\!  \sum_m (l(\tmmathbf{s}, \tmmathbf{c}^{(m\!)}\!)\!-\!b\!-\!b(\tmmathbf{s})) \frac{\partial \log q_\phi (\tmmathbf{c}^{(m\!)} | \tmmathbf{s} \nobracket)}{\partial \phi}
\end{equation}
which is basically a likelihood-ratio estimator.

\section{Forced-attention Sentence Compression}
In neural variational inference, the effectiveness of training largely depends on the quality of the inference network gradient estimator.  
Although we introduce a biased estimator by using pointer networks, it is still very difficult for the compression model to generate reasonable natural language sentences at the early stage of learning, which results in high-variance for the gradient estimator. 
Here, we introduce our supervised forced-attention sentence compression (FSC) model to teach the compression model to generate coherent compressed sentences.

Neither directly replicating the pointer network of ASC model, nor using a typical sequence-to-sequence model, 
the FSC model employs a force-attention strategy (Figure \ref{fig:fsc}) that encourages the compressor to select words appearing in the source sentence but keeps the original full output vocabulary $V$.
The force-attention strategy is basically a combined pointer network that chooses whether to select a word from the source sentence $\tmmathbf{s}$ or to predict a word from $V$ at each recurrent state.
Hence, the combined pointer network learns to copy the source words while predicting the word sequences of compressions.
By sharing the pointer networks between the ASC and FSC model, the biased estimator obtains further positive biases by training on a small set of labelled source-compression pairs.

Here, the FSC model makes use of the compression model (Eq. \ref{eq:bi} to \ref{eq:u}) in the ASC model,
\begin{align}
  \alpha_j  = & \tmop{softmax} ( \tmmathbf{u}_j),
\end{align}
where $\alpha_j(i)$, $i \in ( 1, \ldots, |\tmmathbf{s}|)$ denotes the probability of
selecting $\tmmathbf{s}_i$ as the prediction for $\tmmathbf{c}_j$.

On the basis of the pointer network, we further introduce the probability of predicting $\tmmathbf{c}_j$ that is selected from the full vocabulary,
\begin{equation}
  \beta_j  =  \tmop{softmax} (  \tmmathbf{W}
  \tmmathbf{h}_j^c), \hspace{5pt} 
\end{equation}
where $\beta_j(w) , w \in ( 1, \ldots, | V |) $ denotes the probability of selecting the $w$th from $V$ as the prediction for  $\tmmathbf{c}_j$.
To combine these two probabilities in the RNN, we define a selection factor $\tmmathbf{t}$ for each state output, which computes the semantic similarities between the current state and the attention vector,
\vspace{-0.3em}
\begin{align}
    \tmmathbf{\eta}_j  =   \sum\nolimits_i^{|\tmmathbf{s}|} \alpha_j(i) \tmmathbf{h}^e_i  \\
    \tmmathbf{t}_j =  \sigma(\tmmathbf{\eta}_j^T \tmmathbf{M} \tmmathbf{h}^c_j ).
\end{align}
Hence, the probability distribution over compressed words is defined as,
\begin{align}
  & p ( \tmmathbf{c}_j | \tmmathbf{c}_{1:j-1}, \tmmathbf{s} )   
\!\!  = \!\! \left\{\!\!\! \begin{array}{ll}
     \tmmathbf{t}_j \alpha_j(i) + (1-\tmmathbf{t}_j)  \beta_j(\tmmathbf{c}_j), & \!\!  \tmmathbf{c}_j\!=\!\tmmathbf{s}_i\\
     (1-\tmmathbf{t}_j)  \beta_j(\tmmathbf{c}_j), & \!\!  \tmmathbf{c}_j \!\nin\! \tmmathbf{s}
   \end{array} \right. 
\end{align}
Essentially, the FSC model is the extended compression model of ASC by incorporating the pointer network with a softmax output layer over the full vocabulary. So we employ $\phi$ to denote the parameters of the FSC model $p_{\phi} (\tmmathbf{c} | \tmmathbf{s} )$, which covers the parameters of the variational distribution $q_{\phi} (\tmmathbf{c} | \tmmathbf{s} )$.

\section{Semi-supervised Training}
As the auto-encoding sentence compression (ASC) model grants the ability to make use of an unlabelled dataset, we explore a semi-supervised training framework for the ASC and FSC models. In this scenario we have a labelled dataset that contains source-compression parallel sentences, $(\tmmathbf{s}, \tmmathbf{c}) \in \mathbbm{L}$, and an unlabelled dataset that contains only source sentences $\tmmathbf{s} \in \mathbbm{U}$. 
The FSC model is trained on $\mathbbm{L}$ so that we are able to learn the compression model by maximising the log-probability,
\begin{equation}
  F = \sum_{( \tmmathbf{c}, \tmmathbf{s}) \in \mathbbm{L}} \log p_{\phi} ( \tmmathbf{c} | \tmmathbf{s} ).
\end{equation}
While the ASC model is trained on $\mathbbm{U}$, where we maximise the modified variational lower bound,
\begin{align}
  L\!\! = \!\! \sum_{\tmmathbf{s} \in
  \mathbbm{U}} ( \mathbbm{E} _{q_\phi(\tmmathbf{c}|\tmmathbf{s})}[\log p_\theta(\tmmathbf{s}|\tmmathbf{c})] \!\! - \!\! \lambda D_{KL}[ q_\phi(\tmmathbf{c}|\tmmathbf{s})|| p(\tmmathbf{c})]).
  \label{eq:un}
\end{align}
The joint objective function of the semi-supervised learning is,
\begin{align}
 \nonumber J \!\! = \!\! &      \sum_{\tmmathbf{s} \in
  \mathbbm{U}} ( \mathbbm{E} _{q_\phi(\tmmathbf{c}|\tmmathbf{s})}[\log p_\theta(\tmmathbf{s}|\tmmathbf{c})] \!\! - \!\! \lambda D_{KL}[ q_\phi(\tmmathbf{c}|\tmmathbf{s})|| p(\tmmathbf{c})]) \\
   &   + \sum_{( \tmmathbf{c}, \tmmathbf{s}) \in \mathbbm{L}} \log p_{\phi} (\tmmathbf{c} | \tmmathbf{s} ) .
\end{align}
Hence, the pointer network is trained on both unlabelled data, $\mathbbm{U}$, and labelled data, $\mathbbm{L}$, by a mixed criterion
of REINFORCE and cross-entropy.

\section{Related Work}

As one of the typical sequence-to-sequence tasks, sentence-level summarisation has been explored by a series of discriminative encoder-decoder neural models. 
\newcite{filippova2015sentence} carries out extractive summarisation via deletion with LSTMs, while \newcite{rush2015neural} applies a convolutional encoder and an attentional feed-forward decoder to generate abstractive summarises, which provides the benchmark for the \textit{Gigaword} dataset. \newcite{s2s2016} further improves the performance by exploring multiple variants of RNN encoder-decoder models. 
The recent works \newcite{gulcehre2016pointing}, \newcite{ling2016latent}, \newcite{s2s2016} and \newcite{gu2016incorporating} also apply the similar idea of combining pointer networks and softmax output.
However, different from all these discriminative models above, we explore generative models for sentence compression.
Instead of training the discriminative model on a big labelled dataset, our original intuition of introducing a combined pointer networks is to bridge the unsupervised generative model (ASC) and supervised model (FSC) so that we could utilise a large additional dataset, either labelled or unlabelled, to boost the compression performance.  
\newcite{dai2015semi} also explored semi-supervised sequence learning, but in a pure deterministic model focused on learning better vector representations.

Recently variational auto-encoders have been applied in a variety of fields as deep generative models. 
In computer vision \newcite{kingma2013auto}, \newcite{rezende2014stochastic}, and \newcite{gregor2015draw} have demonstrated strong performance on the task of image generation and \newcite{eslami2016attend} proposed variable-sized variational auto-encoders to identify multiple objects in images.
While in natural language processing, there are variants of VAEs on modelling  documents \cite{miao2015neural}, sentences \cite{bowman2015generating} and discovery of relations \cite{marcheggiani2016discrete}.
Apart from the typical initiations of VAEs, there are also a series of works that employs generative models for supervised learning tasks. 
For instance, \newcite{ba2014multiple} learns visual attention for multiple objects by optimising a variational lower bound, \newcite{kingma2014semi} implements a semi-supervised framework for image classification and \newcite{miao2015neural} applies a conditional variational approximation in the task of factoid question answering. \newcite{dyer2016recurrent} proposes a generative model that explicitly extracts syntactic relationships among words and phrases which further supports the argument that generative models can be a statistically efficient method for learning neural networks from small data.

\section{Experiments}
\subsection{Dataset \& Setup} 
We evaluate the proposed models on the standard \textit{Gigaword}\footnote{https://catalog.ldc.upenn.edu/LDC2012T21} sentence compression dataset. This dataset was generated by pairing the headline of each article with its first sentence to create a source-compression pair. 
\newcite{rush2015neural} provided scripts\footnote{https://github.com/facebook/NAMAS} to filter out outliers, resulting in roughly 3.8M training pairs, a 400K validation set, and a 400K test set. 
In the following experiments all models are trained on the training set with different data sizes\footnote{The hyperparameters where tuned on the validation set to maximise the perplexity of the summaries rather than the reconstructed source sentences.} and tested on a 2K subset, which is identical to the test set used by \newcite{rush2015neural} and \newcite{s2s2016}.
We decode the sentences by $k=5$ Beam search and test with full-length Rouge score.

For the ASC and FSC models, we use 256 for the dimension of both hidden units and lookup tables. In the ASC model, we apply a 3-layer bidirectional RNN with skip connections as the encoder, a 3-layer RNN pointer network with skip connections as the compressor, and a 1-layer vanilla RNN with soft attention as the decoder. 
The language model prior is trained on the article sentences of the full training set using a 3-layer vanilla RNN with 0.5 dropout.
To lower the computational cost, we apply different vocabulary sizes for encoder and compressor (119,506 and 68,897) which corresponds to the settings of \newcite{rush2015neural}. Specifically, the vocabulary of the decoder is filtered by taking the most frequent 10,000 words from the vocabulary of the encoder, where the rest of the words are tagged as `\textit{<unk>}'. 
In further consideration of efficiency, we use only one sample for the gradient estimator.
We optimise the model by Adam \cite{KingmaB14} with a 0.0002 learning rate and 64 sentences per batch. The model converges in 5 epochs.
Except for the pre-trained language model, we do not use dropout or embedding initialisation for ASC and FSC models.

\begin{table*}[!tb]
\centering
\footnotesize
\addtolength{\tabcolsep}{-0.5pt}
\begin{tabular}{c|cc|ccc|ccc|ccc}
    \toprule
    \multirow{2}{*}{Model} &\multicolumn{2}{c|}{Training Data}   &\multicolumn{3}{c|}{Recall}     &\multicolumn{3}{c|}{Precision}     &\multicolumn{3}{c}{F-1}\\ 
    \cline{2-12}
   &Labelled  &Unlabelled  &R-1 &R-2 &R-L &R-1 &R-2 &R-L &R-1 &R-2 &R-L\\
   \hline
    FSC &500K &-       &\textbf{30.817}	&\textbf{10.861}	&\textbf{28.263}	&22.357	&7.998	&20.520	&23.415	&8.156	&21.468\\
ASC+FSC$_1$ &500K &500K &29.117	&10.643	&26.811	&28.558	&10.575	&26.344	&26.987	&9.741	&24.874 \\
ASC+FSC$_2$ &500K &3.8M    &28.236	&10.359	&26.218	&\textbf{30.112}	&\textbf{11.131}	&\textbf{27.896}	&\textbf{27.453}	&\textbf{9.902}	&\textbf{25.452}\\   
 \hline            
    FSC &1M &- &\textbf{30.889}	&\textbf{11.645}	&\textbf{28.257}	&27.169	&10.266	&24.916	&26.984	&10.028	&24.711 \\ 
ASC+FSC$_1$  &1M &1M &30.490	&11.443	&28.097	&28.109	&10.799	&25.943	&27.258	&10.189	&25.148 \\ 
ASC+FSC$_2$  &1M &3.8M  &29.034	 &10.780	&26.801	&\textbf{31.037}	&\textbf{11.521}	&\textbf{28.658}	&\textbf{28.336}	&\textbf{10.313}	&\textbf{26.145}\\
   \hline
    FSC     &3.8M  &-       &\textbf{30.112}	&12.436	&\textbf{27.889}	&34.135	&13.813	&31.704	&30.225	&12.258	&28.035  \\
    ASC+FSC$_1$ &3.8M  &3.8M    &29.946	&\textbf{12.558}	&27.805	&\textbf{35.538}	&\textbf{14.699}	&\textbf{32.972}	&\textbf{30.568}	&\textbf{12.553}	&\textbf{28.366}  \\
    \bottomrule
  \end{tabular}
  \caption{Extractive Summarisation Performance. (1) The extractive summaries of these models are decoded by the pointer network (i.e the shared component of the ASC and FSC models).
(2) R-1, R-2 and R-L represent the Rouge-1, Rouge-2 and Rouge-L score respectively.}
 \label{tb:ex}
\end{table*}

\subsection{Extractive Summarisation}
The first set of experiments evaluate the models on extractive summarisation.
Here, we denote the joint models by ASC+FSC$_1$ and ASC+FSC$_2$ where ASC is trained on unlabelled data and FSC is trained on labelled data. 
The ASC+FSC$_1$ model employs equivalent sized labelled and unlabelled datasets, where the article sentences of the unlabelled data are the same article sentences in the labelled data, so there is no additional unlabelled data applied in this case. 
The ASC+FSC$_2$ model employs the full unlabelled dataset in addition to the existing labelled dataset, which is the true semi-supervised setting. 

Table \ref{tb:ex} presents the test Rouge score on extractive compression.
We can see that the ASC+FSC$_1$ model achieves significant improvements on F-1 scores when compared to the supervised FSC model only trained on labelled data.
Moreover, fixing the labelled data size, the ASC+FSC$_2$ model achieves better performance by using additional unlabelled data than the ASC+FSC$_1$ model, which means the semi-supervised learning works in this scenario.
Interestingly, learning on the unlabelled data largely increases the precisions (though the recalls do not benefit from it) which leads to significant improvements on the F-1 Rouge scores. And surprisingly, the extractive ASC+FSC$_1$ model trained on full labelled data outperforms the abstractive NABS \cite{rush2015neural} baseline model (in Table \ref{tb:rouge}).

\subsection{Abstractive Summarisation}
The second set of experiments evaluate performance on abstractive summarisation (Table \ref{tb:ab}).
Consistently, we see that adding the generative objective to the discriminative model (ASC+FSC$_1$) results in a significant boost on all the Rouge scores, while employing extra unlabelled data increase performance further (ASC+FSC$_2$).
This validates the effectiveness of transferring the knowledge learned on unlabelled data to the supervised abstractive summarisation.

\begin{figure}[!t] 
  \centering
	\includegraphics[width=3.1in]{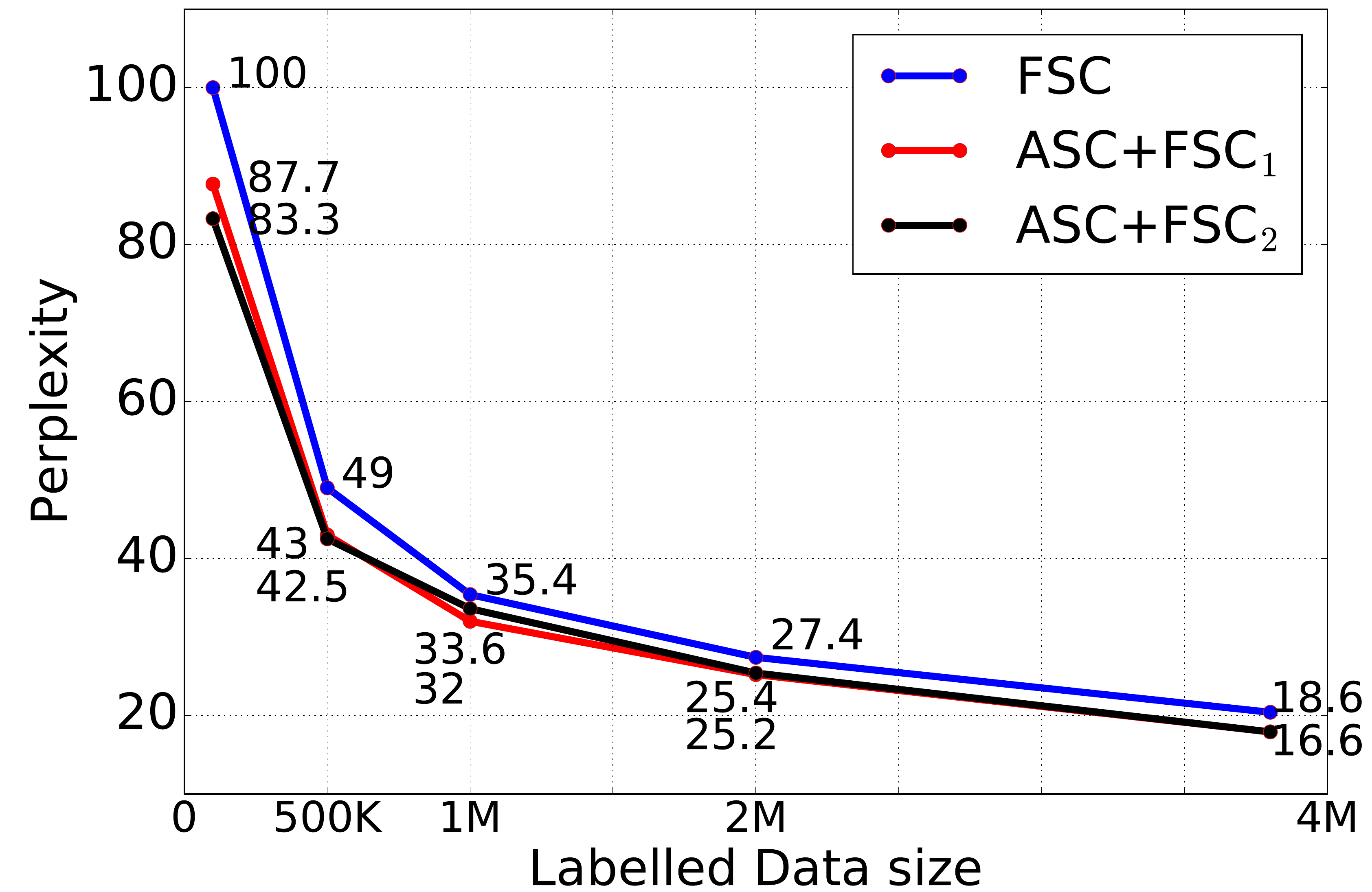} 
  \caption{Perplexity on validation dataset.}
  \label{fig:ppx}
  \vspace{-1em}
\end{figure}

In Figure \ref{fig:ppx}, we present the validation perplexity to compare the abilities of the three models to learn the compression languages.  
The ASC+FSC$_1$(red) employs the same dataset for unlabelled and labelled training, while the ASC+FSC$_2$(black) employs the full unlabelled dataset.
Here, the joint ASC+FSC$_1$ model obtains better perplexities than the single discriminative FSC model, but there is not much difference between ASC+FSC$_1$ and ASC+FSC$_2$ when the size of the labelled dataset grows. 
From the perspective of language modelling, the generative ASC model indeed helps the discriminative model learn to generate good summary sentences.
Table \ref{tb:ppx} displays the validation perplexities of the benchmark models, where the joint ASC+FSC$_1$ model trained on the full labelled and unlabelled datasets performs the best on modelling compression languages.

\begin{table*}[!tb]
\centering
\footnotesize
\addtolength{\tabcolsep}{-0.5pt}
\begin{tabular}{c|cc|ccc|ccc|ccc}
    \toprule
    \multirow{2}{*}{Model} &\multicolumn{2}{c|}{Training Data}   &\multicolumn{3}{c|}{Recall}     &\multicolumn{3}{c}{Precision} &\multicolumn{3}{c}{F-1} \\ 
    \cline{2-12}
   &Labelled  &Unlabelled  &R-1 &R-2 &R-L &R-1 &R-2 &R-L &R-1 &R-2 &R-L \\
   \hline
    FSC     &500K &-        &27.147	&10.039	&25.197	&33.781	&13.019	&31.288	&29.074	&10.842	&26.955   \\
    ASC+FSC$_1$ &500K &500K     &27.067	&10.717	&25.239	&33.893	&13.678	&31.585	&29.027	&11.461	&27.072  \\
    ASC+FSC$_2$ &500K &3.8M     &\textbf{27.662}	&\textbf{11.102}	&\textbf{25.703}	&\textbf{35.756}	&\textbf{14.537}	&\textbf{33.212}	&\textbf{30.140}	&\textbf{12.051}	&\textbf{27.99}	 \\
   \hline
    FSC     &1M &-          &28.521	&11.308	&26.478	&33.132	&13.422	&30.741	&29.580	&11.807	&27.439  \\
    ASC+FSC$_1$ &1M &1M         &28.333	&11.814	&26.367	&35.860	&\textbf{15.243}	&33.306	&30.569	&12.743	&28.431  \\
    ASC+FSC$_2$ &1M &3.8M       &\textbf{29.017}	&\textbf{12.007}	&\textbf{27.067}	&\textbf{36.128}	&14.988	&\textbf{33.626}	&\textbf{31.089}	&\textbf{12.785}	&\textbf{28.967}     \\
   \hline
    FSC     &3.8M  &-       &31.148	&13.553	&28.954	&36.917	&16.127	&34.405	&32.327	&14.000	&30.087  \\
    ASC+FSC$_1$ &3.8M  &3.8M    &\textbf{32.385}	&\textbf{15.155}	&\textbf{30.246}	&\textbf{39.224}	&\textbf{18.382}	&\textbf{36.662}	&\textbf{34.156}	&\textbf{15.935}	&\textbf{31.915}   \\
    \bottomrule
  \end{tabular}
  \caption[table]{Abstractive Summarisation Performance. The abstractive summaries of these models are decoded by the combined pointer network (i.e. the shared pointer network together with the softmax output layer over the full vocabulary). }
 \label{tb:ab}
 \vspace{-1em}
\end{table*}

Table \ref{tb:rouge} compares the test Rouge score on abstractive summarisation. 
Encouragingly, the semi-supervised model ASC+FSC$_2$ outperforms the baseline model NABS when trained on 500K supervised pairs, which is only about an eighth of the supervised data.
In \newcite{s2s2016}, the authors exploit the full limits of discriminative RNN encoder-decoder models by incorporating a sampled softmax, expanded vocabulary, additional lexical features, and combined pointer networks\footnote{The idea of the combined pointer networks is similar to the FSC model, but the implementations are slightly different.}, which yields the best performance listed in Table \ref{tb:rouge}. 
However, when all the data is employed with the mixed objective ASC+FSC$_1$ model, the result is significantly better than this previous state-of-the-art.
As the semi-supervised ASC+FSC$_2$ model can be trained on unlimited unlabelled data, there is still significant space left for further performance improvements.

Table \ref{tb:sample} presents the examples of the compression sentences 
decoded by the joint model ASC+FSC$_1$  and the FSC model trained on the full dataset.

\begin{table}[!t]
	\centering
	\footnotesize
	\addtolength{\tabcolsep}{-3.5pt}
	\begin{tabular}{@{}llcc@{}}
    		\toprule
    		\multicolumn{2}{@{}l}{Model}  & Labelled Data & Perplexity \\
		\midrule
		&Bag-of-Word (BoW)&3.8M &43.6\\
		&Convolutional (TDNN) &3.8M &35.9\\
		&Attention-Based (NABS)  &3.8M &27.1\\
		&\cite{rush2015neural}&\\
    		\midrule
    		& Forced-Attention (FSC) 	&3.8M & 18.6 \\
    		& Auto-encoding (ASC+FSC$_1$)&3.8M  &  16.6 \\
    		\bottomrule
  	\end{tabular}
  	\captionof{table}{Comparison on validation perplexity. BoW, TDNN and NABS are the baseline neural compression models with different encoders in \protect\newcite{rush2015neural}}
 	\label{tb:ppx}
\end{table}

\begin{table}[!t] 
	\centering
	\footnotesize
	\addtolength{\tabcolsep}{-2.5pt}
	\begin{tabular}{cc|ccc}
    		\toprule
    		Model & Labelled Data & R-1 & R-2 & R-L \\
		\hline
	  \cite{rush2015neural} &3.8M &29.78	&11.89	&26.97 \\
		\cite{s2s2016} &3.8M & 33.17	&\textbf{16.02}	&30.98 \\
    		\hline
    		 ASC + FSC$_2$ &500K &30.14	&12.05	&27.99 \\
			 ASC + FSC$_2$ &1M &31.09	&12.79	&28.97\\
    		 ASC + FSC$_1$ &3.8M & \textbf{34.17}	&15.94	&\textbf{31.92} \\
    		\bottomrule
  	\end{tabular}
  	\caption{Comparison on test Rouge scores}
  	 \vspace{-2em}
 	\label{tb:rouge}
\end{table}

\section{Discussion}
From the perspective of generative models, a significant contribution of our work is a process for reducing variance for discrete sampling-based variational inference. 
The first step is to introduce two baselines in the control variates method due to the fact that the reparameterisation trick is not applicable for discrete latent variables. 
However it is the second step of using a pointer network as the biased estimator that makes the key contribution. This results in a much smaller state space, bounded by the length of the source sentence (mostly between 20 and 50 tokens), compared to the full vocabulary. 
The final step is to apply the FSC model to transfer the knowledge learned from the supervised data to the pointer network.
This further reduces the sampling variance by acting as a sort of bootstrap or constraint on the unsupervised latent space which could encode almost anything but which thus becomes biased towards matching the supervised distribution.
By using these variance reduction methods, the ASC model is able to carry out effective variational inference for the latent language model so that it learns to summarise the sentences from the large unlabelled training data.

In a different vein, according to the reinforcement learning interpretation of sequence level training \cite{ranzato2015sequence}, the compression model of the ASC model acts as an agent which iteratively generates words (takes actions) to compose the compression sentence and the reconstruction model acts as the reward function evaluating the quality of the compressed sentence which is provided as a reward signal. 
\newcite{ranzato2015sequence} presents a thorough empirical evaluation on three different NLP tasks by using additional sequence-level reward (BLEU and Rouge-2) to train the models. 
In the context of this paper, we apply a variational lower bound (mixed reconstruction error and KL divergence regularisation) instead of the explicit Rouge score.
Thus the ASC model is granted the ability to explore unlimited unlabelled data resources.
In addition we introduce a supervised FSC model to teach the compression model to generate stable sequences instead of starting with a random policy.
In this case, the pointer network that bridges the supervised and unsupervised model is trained by a mixed criterion of REINFORCE and cross-entropy in an incremental learning framework.
Eventually, according to the experimental results, the joint ASC and FSC model is able to learn a robust compression model by exploring both labelled and unlabelled data, which outperforms the other single discriminative compression models that are only trained by cross-entropy reward signal.

\section{Conclusion}
In this paper we have introduced a generative model for jointly modelling pairs of sequences and evaluated its efficacy on the task of sentence compression. The variational auto-encoding framework provided an effective inference algorithm for this approach and also allowed us to explore combinations of discriminative (FSC) and generative (ASC) compression models. The evaluation results show that supervised training of the combination of these models improves upon the state-of-the-art performance for the Gigaword  compression dataset. When we train the supervised FSC model on a small amount of labelled data and the unsupervised ASC model on a large set of unlabelled data the combined model is able to outperform previously reported benchmarks trained on a great deal more supervised data. These results demonstrate that we are able to model language as a discrete latent variable in a variational auto-encoding framework and that the resultant generative model is able to effectively exploit both supervised and unsupervised data in sequence-to-sequence tasks.

\begin{table}[H] 
	\centering
	\scriptsize
	\addtolength{\tabcolsep}{-3.5pt}
	\begin{tabular}{m{13pt}|m{210pt}}
    	\toprule
\color{blue!60!black}\textbf{src} \vspace{20pt} &\color{blue!60!black}the sri lankan government on wednesday announced the closure of government schools with immediate effect as a military campaign against tamil separatists escalated in the north of the country . \\
\color{blue!60!black}\textbf{ref} \vspace{2pt}&\color{blue!60!black}sri lanka closes schools as war escalates \\
[0.03in]
\textbf{asc$_a$} &sri lanka closes government schools \\
\textbf{asc$_e$} &sri lankan government closure schools escalated \\
\textbf{fsc$_a$} &sri lankan government closure with tamil rebels closure \\ 
		\hline
\color{blue!60!black}\textbf{src} \vspace{10pt}&\color{blue!60!black}factory orders for manufactured goods rose \#.\# percent in september , the commerce department said here thursday .\\
\color{blue!60!black}\textbf{ref} \vspace{2pt}&\color{blue!60!black}us september factory orders up \#.\# percent \\ 
[0.03in]
\textbf{asc$_a$} &us factory orders up \#.\# percent in september \\
\textbf{asc$_e$} &factory orders rose \#.\# percent in september \\
\textbf{fsc$_a$} &factory orders \#.\# percent in september \Bstrut\\ 
		\hline
\color{blue!60!black}\textbf{src} \vspace{20pt}&\color{blue!60!black}hong kong signed a breakthrough air services agreement with the united states on friday that will allow us airlines to carry freight to asian destinations via the territory . \\
\color{blue!60!black}\textbf{ref} \vspace{3pt}&\color{blue!60!black}hong kong us sign breakthrough aviation pact \\
[0.03in]
\textbf{asc$_a$} &us hong kong sign air services agreement \\
\textbf{asc$_e$} &hong kong signed air services agreement with united states \\
\textbf{fsc$_a$} &hong kong signed air services pact with united states \Bstrut\\ 
		\hline
\color{blue!60!black}\textbf{src} \vspace{20pt}&\color{blue!60!black}a swedish un soldier in bosnia was shot and killed by a stray bullet on tuesday in an incident authorities are calling an accident , military officials in stockholm said tuesday .\\
\color{blue!60!black}\textbf{ref} \vspace{2pt}&\color{blue!60!black}swedish un soldier in bosnia killed by stray bullet \\
[0.03in]
\textbf{asc$_a$} &swedish un soldier killed in bosnia \\
\textbf{asc$_e$} &swedish un soldier shot and killed \\
\textbf{fsc$_a$} &swedish soldier shot and killed in bosnia \Bstrut\\ 
		\hline
\color{blue!60!black}\textbf{src} \vspace{10pt}&\color{blue!60!black}tea scores on the fourth day of the second test between australia and pakistan here monday .\\
\color{blue!60!black}\textbf{ref} \vspace{2pt}&\color{blue!60!black}australia vs pakistan tea scorecard \\
[0.03in]
\textbf{asc$_a$} &australia v pakistan tea scores \\
\textbf{asc$_e$} &australia tea scores \\
\textbf{fsc$_a$} &tea scores on \#th day of \#nd test \Bstrut\\ 
		\hline
\color{blue!60!black}\textbf{src} \vspace{20pt}&\color{blue!60!black}india won the toss and chose to bat on the opening day in the opening test against west indies at the antigua recreation ground on friday .\\
\color{blue!60!black}\textbf{ref} \vspace{2pt}&\color{blue!60!black}india win toss and elect to bat in first test \\
[0.03in]
\textbf{asc$_a$} &india win toss and bat against west indies \\
\textbf{asc$_e$} &india won toss on opening day against west indies \\
\textbf{fsc$_a$} &india chose to bat on opening day against west indies \Bstrut\\ 
		\hline
\color{blue!60!black}\textbf{src} \vspace{20pt}&\color{blue!60!black}a powerful bomb exploded outside a navy  base near the sri lankan capital colombo tuesday , seriously wounding at least one person , military officials said .\\
\color{blue!60!black}\textbf{ref} \vspace{2pt}&\color{blue!60!black}bomb attack outside srilanka navy base \\
[0.03in]
\textbf{asc$_a$} &bomb explodes outside sri lanka navy base \\
\textbf{asc$_e$} &bomb outside sri lankan navy base wounding one \\
\textbf{fsc$_a$} &bomb exploded outside sri lankan navy base \Bstrut\\ 
		\hline
\color{blue!60!black}\textbf{src} \vspace{20pt}&\color{blue!60!black} press freedom in algeria remains at risk despite the release on wednesday of prominent newspaper editor mohamed <unk> after a two-year prison sentence , human rights organizations said .\\
\color{blue!60!black} \textbf{ref} \vspace{10pt}&\color{blue!60!black} algerian press freedom at risk despite editor 's release <unk> picture\\
[0.03in]
\textbf{asc$_a$}&algeria press freedom remains at risk\\
\textbf{asc$_e$} &algeria press freedom remains at risk\\
\textbf{fsc$_a$} &press freedom in algeria at risk\Bstrut\\
    	\bottomrule
  	\end{tabular}
  	\caption{Examples of the compression sentences. \textbf{src} and \textbf{ref} are the source and reference sentences provided in the test set. \textbf{asc$_a$} and \textbf{asc$_e$} are the abstractive and extractive compression sentences decoded by the joint model ASC+FSC$_1$, and \textbf{fsc$_a$} denotes the abstractive compression obtained by the FSC model. }
 	\label{tb:sample}
\end{table}

\bibliography{emnlp2016}
\bibliographystyle{emnlp2016}

\end{document}